\documentclass[journal]{IEEEtran}
\ifCLASSINFOpdf
\else
\fi

\usepackage{xcolor}
\usepackage{float}
\usepackage{graphicx}
\usepackage{multicol, blindtext}
\usepackage{multirow}
\usepackage{adjustbox}
\usepackage{amsfonts}
\usepackage{subcaption}
\usepackage{amsmath}
\usepackage{cleveref}
\begin{document}
%

\title{Deep Learning and Handheld Augmented Reality Based System for Optimal Data Collection in Fault Diagnostics Domain}
%
%
%

\author{Ryan~Nguyen~and~Rahul~Rai~\thanks{R. Nguyen, Ph.D. Student, Department
of Automotive Engineering, Clemson University, Clemson,
SC, 29634 USA e-mail: rnguye2@clemson.edu.}
\thanks{R. Rai, Corresponding Author, Dean's Distinguished Professor, Department
of Automotive Engineering, Clemson University, Clemson,
SC, 29634 USA e-mail: rrai@clemson.edu.}}

%
%

\markboth{IEEE TRANSACTIONS ON HUMAN-MACHINE SYSTEMS}%
{Nguyen and Rai: Deep Learning and Handheld Augmented Reality Based System for Optimal Data Collection in Fault Diagnostics Domain}

\maketitle
\emph{This work has been submitted to the IEEE for possible publication. Copyright may be transferred without notice, after which this version may no longer be accessible.}

\begin{abstract}

 Compared to current AI or robotic systems, humans navigate their environment with ease, making tasks such as data collection trivial. However, humans find it harder to model complex relationships hidden in the data. AI systems, especially deep learning (DL) algorithms, impressively capture those complex relationships. Symbiotically coupling humans and computational machines' strengths can simultaneously minimize the collected data required and build complex input-to-output mapping models. This paper enables this coupling by presenting a novel human-machine interaction framework to perform fault diagnostics with minimal data. Collecting data for diagnosing faults for complex systems is difficult and time-consuming. Minimizing the required data will increase the practicability of data-driven models in diagnosing faults. The framework provides instructions to a human user to collect data that mitigates the difference between the data used to train and test the fault diagnostics model. The framework is composed of three components: (1) a reinforcement learning algorithm for data collection to develop a training dataset, (2) a deep learning algorithm for diagnosing faults, and (3) a handheld augmented reality application for data collection for testing data. The proposed framework has provided above 100\% precision and recall on a novel dataset with only one instance of each fault condition. Additionally, a usability study was conducted to gauge the user experience of the handheld augmented reality application, and all users were able to follow the provided steps. 

\end{abstract}
\section{Introduction}







A proper fault diagnosis can provide a system operator with helpful information about the required maintenance on a system. Damage from faulty system behavior can cause long downtimes and high operations and maintenance costs. Capturing the system fault earlier can minimize the damage from a non-optimal operating condition \cite{9486591}. Fault diagnostics can be performed either using physics-based methods or data-driven methods. Physics-based methods are predicated on observing the system behavior or deriving interactions from first principles \cite{parlos2000multi}. The faults of complex systems are governed by complex relationships that are difficult to model using a physics-based approach. Additionally, due to limited knowledge of the underlying physics, system modelers often make simplifying assumptions and construct coarse models that describe system behavior imperfectly. These drawbacks of physics-based modeling approaches have lent credibility to data-driven methods and made them more attractive \cite{kavulya2012failure}. 

Data-driven methods do not require knowledge of the underlying physics; instead, the modeling process iterates over a dataset and updates the model based on how well it predicts the outputs from input data. Therefore, the modeling process is automated and can be widely applied. Nevertheless, high-performing data-driven methods, such as deep learning (DL) approaches, require a large amount of data, time, and computation power to train and diagnose faults accurately. When abundant data is available, and when time or computation power are not limiting factors, DL has been able to diagnose faults in bearings \cite{zhang2020deep}, rotor shafts \cite{li2019deep}, and drive trains \cite{bach2018deep}. Major limitations of existing purely data-driven statistical “black box” methods include \cite{sahu2020artificial}: (a) their inability to generalize beyond their initial set of training data, (b) their agnostic view of underlying physics, resulting in model outputs that lack scientific coherency with the known laws of physics, and (c) their “data-hungry” nature that precludes them from being used in scientific problems and applications with limited or sparse data.

The limitations of physics-based and data-driven methods can be addressed by designing efficient and symbiotic human-machine interaction (HMI) for fault diagnostics. Recent domains that benefit from HMI symbiosis are user experience (UX) design \cite{hassenzahl2013user}, digital interfaces \cite{downs2005using}, and devices for the Internet of Things (IoT) \cite{nuamah2017human}. Stephanidis et al. \cite{stephanidis2019seven} elevate the importance of human-AI symbiosis; it is the first of the seven HMI grand challenges introduced. Stephanidis et al. \cite{stephanidis2019seven} explore how our synergy with technology is inevitable and stress the importance of developing techniques that will allow us to transition elegantly. Li et al.
\cite{li2020reciprocity} explore modern techniques of human-agent cooperation; the agent is trained to perform a task, and the human performs the task alongside the agent. Both communicate and coordinate laterally to handle tasks in dynamically changing environments, demonstrating the benefit of evolving from interactions to coordination. For human-AI systems, techniques for improving communication include human-centered reinforcement learning and active learning. 



Human-centered reinforcement learning (HCRL) is a reinforcement learning framework that includes a human's feedback in training--the reinforcement learning algorithm's reward is based on the feedback of the human observer. Li et al. \cite{8708686} provide a comprehensive survey of the techniques in the field such as interactive shaping, learning from categorical feedback, and learning from policy feedback and provide future directions such as exploring human-centered deep reinforcement learning and gaining a deeper understanding of human evaluative feedback to measure the efficacy of the feedback provided to the learning algorithm. HCRL suffers from inconsistent human feedback and requires a human presence during training, making the training less automated. 
Like HCRL, active learning algorithms query a user, but instead of using the human's input as part of the reward, the human is used to label necessary data during training. The process minimizes the amount of labeled data required while preserving large unlabeled datasets. Gal et al. \cite{gal2017deep}, Sener et al. \cite{sener2017active}, and Yang et al. \cite{yang2017suggestive} demonstrate active learning's effect on convolutional neural networks' (CNN) performance on benchmark datasets (i.e., MNIST, CIFAR, etc.) and show how the improvements allow for these algorithms to be applied to practical domains such as biomedical imaging analysis \cite{yang2017suggestive}. This online querying develops a relationship with DL and demonstrates the potency of communication. However, although active learning has shown to be effective at training DL architectures, the technique still requires a lot of data and a long training time; additionally, in active learning, the human needs to be present for the system to work optimally.

Due to the inconsistency of human feedback in HCRL and the long training times of active learning, we propose a method that does not rely on human feedback during training and minimizes the dataset size to reduce training time–specifically in the fault diagnostics domain. Our method (Figure \ref{fig:interactions_main}) comprises three components: (1) A reinforcement learning (RL) algorithm, (2) A deep learning (DL) algorithm, and (3) A handheld augmented reality (HAR) application. The RL enables data collection and produces instructions that the HAR application will display–the instructions are followed to develop a training dataset comprised of virtual data that maximizes the difference between fault conditions. This training dataset is provided to the DL to perform the diagnosis--after the DL is fully trained, the final model is saved and deployed on the HAR application. The HAR application displays the instructions discovered by the RL to a human user for collecting testing data that resembles the data in the training dataset--this will minimize any error due to overfitting and maximize the prediction accuracy of the trained DL.


This method minimizes the similarity between classes within a training dataset and provides users with instructions to collect data similar to the training dataset for testing–this reduces the dataset size required to train a DL model, reducing the training time and computational power.  This paper discusses the approach in further detail and offers the following contributions:
\begin{itemize}
    \item Introduces a reward function for an RL algorithm to minimize dataset size and maximize class differentiation.
    \item Measures the precision and recall of a DL model using the dataset collected by the RL algorithm and compares the results to random datasets for a diagnostics problems.
    \item Introduces a HAR application that improves the human-AI relationship by guiding a user to collect data similar to the training dataset.
\end{itemize}


Section \ref{Overview} provides a brief overview of the framework and provides a diagram to improve clarity. Section \ref{ProblemForm} formally introduces the problem. Then, Section \ref{Methodology} provides the methodology for solving the problem by including the network details for the RL algorithm and the DL algorithm.  Next, Section \ref{ProbSet} discusses the example problem set for benchmarking the presented HAR-DL pipeline (HARDLearning). Lastly, \ref{Results}-\ref{Conclusion} provide the results and conclude the paper.

\section{Overview}
\label{Overview}
\begin{figure*}[ht]
    \centering
    \includegraphics[height=8cm]{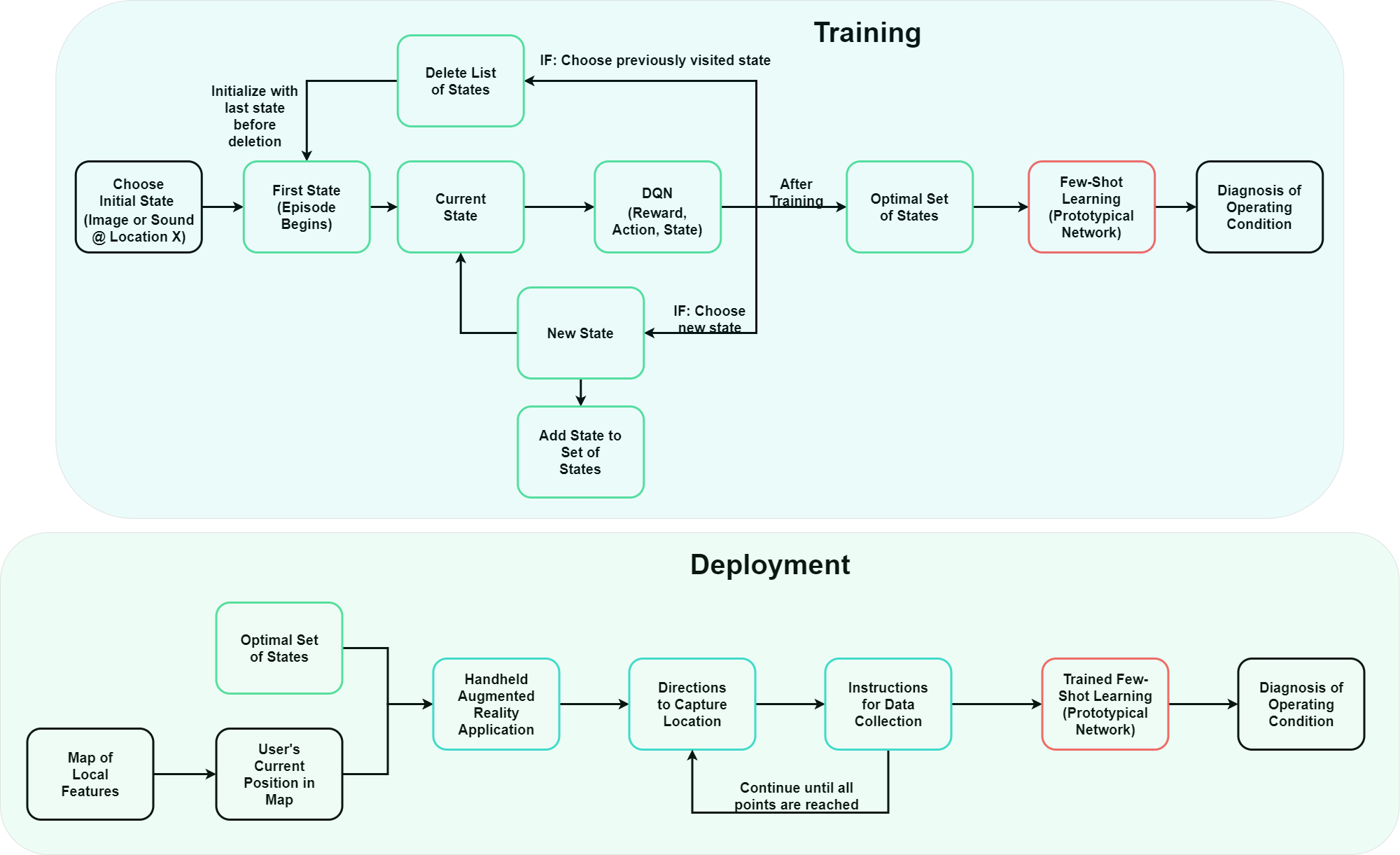}
    \caption{The main components of the proposed human-machine interaction framework. The first half of the diagram refers to the training portion of the framework where the deep Q-Network (DQN) and the few-shot learning (FSL) algorithm, prototypical network, are trained. The trained DQN outputs the optimal set of states which is used to guide users to the corresponding real-world locations in the deployment phase of the framework. Additionally, the trained prototypical network is used to map the collected information to the current operating condition.}
    \label{fig:interactions_main}
\end{figure*}

This paper aims to develop a framework that provides a user with data collection instructions to optimize the performance of a fully trained few-shot learning (FSL) classification model for diagnostics (Figure \ref{fig:interactions_main}). We hypothesize that there exists a dataset that optimizes the accuracy of the FSL model and that this dataset consists of data collected at specific locations around the system being monitored that represent the greatest differentiation between the operating conditions. These points are explored virtually using a deep Q-network (DQN) \cite{mnih2013playing} which is initialized with a randomly selected location, collecting either visual data modeled by a 3D model or auditory data in the form of wave files of the real-world system prior to testing (explained in Section \ref{ProbSet}). DQN chooses the following location and data type to collect. This process is repeated to append to the dataset collected for that episode; if that state has not been visited before, that new location will be added to the set of locations visited, and DQN will continue collecting data. If DQN reaches a location that has been visited, then the episode ends, and if this is not the last episode, then the list is deleted, and the last data point of the deleted list becomes the first data point of the new episode. This is repeated until DQN finishes training. The dataset developed by the visited points of the last training episode is used to train the FSL model. The FSL model uses the data from those locations to predict the operating condition. 
Once the FSL model is fully trained, it is saved and deployed on a HAR application.

The set of points visited by DQN for data collection is imported into the HAR application for a user to diagnose a system's operating condition in the real world. This application initializes the user in a map of local features pre-established around the system and provides instructions to navigate the user from their initial position to the locations for data collection. After navigating to a location for data collection, data collection instructions will populate the screen for the user to follow. Once all locations have been visited, and data collection is complete, the data will be provided to the FSL model saved on the application and predict the operating condition. These steps are outlined in Figure \ref{fig:interactions_main} and is revisited in more detail in Section \ref{Methodology}.

 Auditory data is available for the problem set, and eight data collection locations are available. Once the data is collected, the auditory data is reformatted, and the audio recordings are transformed into Mel-spectrogram images. The Mel scale \cite{pedersen1965mel} is a perceptual scale of pitches determined by listeners to be equal in distance from one another. The reference point between this scale and the normal frequency measurement is defined by assigning a perceptual pitch of 1000 Mels to a 1000 Hz tone, 40 dB above the listener's threshold. 

Continuing the preprocessing, both the visual data and Mel-spectrograms were resized to 120 pixels by 160 pixels to reduce the size of the network. The image is normalized by dividing each element in the matrix by 255. The normalization helps prevent exploding gradient issues which leads to improved accuracy. The tensors are organized in the following way: (1) the index of the image in the list, (2) the channels of color, (3) the height, and (4) the width. This arrangement of data is the format expected by the convolution layers, so it is imperative to maintain this organization. The process of transforming the data is explained in detail in Section \ref{ProbSet}.

Lastly, applying HARDLearning to the fan problem set, DQN returned one data collection location and the dataset was able to train the FSL model to predict the operating condition with  $100\%$ precision, and recall, Section \ref{Results} elucidates these findings.

\section{Problem Formulation}
\label{ProblemForm}

This section formally introduces the problems. The first subsection introduces the higher-level problem of the suboptimal human-AI relationship and the second subsection presents the problem of developing an optimal dataset for a DL model.
\subsection{Human-AI Symbiosis}

Symbiotic communication between humans and AI systems is the optimal sequence of interactions between two entities; for problem-solving, symbiosis should achieve the following goals:
\begin{enumerate}
  \item Develop a map, $F: X \longrightarrow Y$ from input data, $X$, to output, $Y$.
  \item Provide the steps, $S_1$, that are required to develop $F$.
  \item Produce a procedure, $S_2$, for collecting the data required to apply $F$.
  \item Display the results for interpretation to achieve a complete understanding.
\end{enumerate}

In human-human communication, symbiosis is achieved through a sequence of steps, $H_h$ ($S_1,S_2 \subset H_h$). Currently, in non-symbiotic human-AI communication, $H_{AI}$, $S_1$ and $S_2$ are not carried out and the DL model, $\hat{F}: X \longrightarrow \hat{Y}$, in $H_{AI}$ might have abundant training data but not the data required to approximate $F$ optimally. The lack of achieving an optimal approximation of $F$ either means $F(X) - \hat{F}(X) > F(X) - \hat{F^*}(X)$ or $t > t^*$ (where $^*$ refers to optimal). $H_h$ solves this problem through communication--each person improves their understanding of $F$ by gaining information from the other person; this process of gaining information from one another is used to develop $S_1$ and will be referred to as $S_0$. $S_0$ is a longer sequence than $S_1$, $S_0$ is all communication between two entities in developing $F$, $S_1$ is the set of steps in $S_0$ that are required to develop $F$, and if a new entity were to be introduced, that entity would only need $S_1$ to develop $F$. If the prediction was optimal in DL, then $S_0$ would not be required, but because the approximation is non-optimal, $S_0$ is imperative for performance. Since $S_0$ is not modeled in $H_{AI}$, the accuracy of the model relies on how well the DL can extrapolate from the data provided rather than cooperating with a human to receive more useful data. Secondly, $H_{AI}$ this leads to an absence of $S_1$. 

These issues make DL impractical for tasks requiring highly reliable models and lacking technical resources to handle long training times. Therefore, the lack of symbiotic communication between humans and AI systems needs to be addressed to maximize accuracy and minimize training time--this is imperative for the future usability of DL systems. 


\subsection{Optimal Data Collection}
The second problem addresses the development of an optimal dataset. An optimal dataset should minimize training time and maximize accuracy for a DL model, $\hat{F}: X \longrightarrow \hat{Y}$. However, the training time, $t$, for a DL model, is directly related to the length of the training set, $n \in \mathbb{N}$; therefore, we frame the problem w.r.t. minimizing $n$. $x \in X$ represents the data collected, and the optimal dataset, $X^*$, should be the dataset that minimizes $n$ and maximizes accuracy. Therefore, the problem is posed as a multi-objective problem that minimizes $\frac{1}{n}\sum^n_{k=0}|y_k - \hat{F}(x_k)|$ and $n$.

\begin{equation}
\begin{array}{rrclcl}
\displaystyle \min_{x_k \in X, n \in \mathbb{N}} & \multicolumn{3}{l}{\frac{1}{n}\sum_{k=1}^n \lvert y_k - \hat{F}(x_k) \rvert} + n\\
\displaystyle
\textrm{s.t.} & n > 0\\
\end{array}
\end{equation}


Addressing optimal data collection is necessary for solving the human-AI symbiosis problem. Our method for solving these two problems is outlined in Section \ref{Methodology}.

\section{Methodology}
This section outlines the overall computation approach and network details for training and testing purposes. HARDLearning proposes improving human-AI communication and helps an agent develop the optimal dataset for a DL model to diagnose faults accurately. 

\label{Methodology}

Modifying $H_{AI}$ helps improve the efficacy of DL models; active learning (AL) methods introduce querying a user to label unlabeled data to help the learning model develop a map. The modification has shown impressive improvements on CNN models \cite{gal2017deep, sener2017active, yang2017suggestive}; however, AL still has long training times due to large training datasets. Thus, we introduce HARDLearning, which modifies $H_{AI}$ by appending versions of $S_0$, $S_1$, and $S_2$ that mimic $S_0$, $S_1$ and $S_2$ of $H_h$ to provide the most useful data for maximizing the accuracy of the approximation of $F$ and reducing the training dataset size to minimize training time. We suggest the optimal modifications drive $H_{AI}$ closer to $H_h$.

HARDLearning is composed of three different phases: (1) an RL algorithm for data collection to develop a training dataset, (2) a DL algorithm for diagnosing faults, and (3) a HAR application for data collection for testing data. The first and second phases are purely computational and add $S_0$ and $S_1$ to $H_{AI}$ and develop $F$, respectively; and the third phase adds $S_2$ to $H_{AI}$ by introducing a communication procedure to instruct a human user to perform test data collection through a HAR application.

\subsection{Training}
The first phase of HARDLearning is training an RL algorithm to provide a DL algorithm with a training dataset. The second phase uses the DL algorithm to develop an approximation of $F$ to perform the fault diagnostics.
\subsubsection{Reinforcement Learning}
The RL algorithm used is a DQN \cite{mnih2013playing}. 
The DQN's states and actions will be defined by a virtual model of a real world system to be monitored. DQN is trained to manipulate the virtual model, the states of DQN will be the locations to collect visual or auditory data which will append to a set of points, $P$, that minimizes the dataset size and maximizes the difference between operating conditions. DQN's actions will be represented by the act of moving from one location to another, DQN can decide to move to an unseen location or a seen location. Visiting a previously visited location will end the episode; whereas, visiting a new location will append data to the dataset being collected during that episode. In terms of the steps for optimal communication, $S_0$ is automated with this process; the human user is providing access to a virtual representation with the modalities mentioned above of the desired system, and DQN can access any of that information and choose the information at the optimal set of points, $P^*$. $P^*$ will be saved and deployed to the HAR application–-directions to those points will be mapped to actions for the user to follow around the real-world model. Secondly, $P^*$ coupled with the type of data collected at those points represents $S_1$ for $H_{AI}$. DQN does not return the classifications–that task is assigned to a second network; the second network is derived from a class of DL algorithms known as few-shot learning (FSL).

The DQN's reward function is composed of three terms. The first term rewards minimizing the average similarity between operating conditions. The second term rewards reducing the maximum class similarity of the average similarity between the data collected for each operating condition. The third term penalizes the amount of data collected. These three terms help DQN find a local minimum that trains the FSL algorithm to differentiate between the operating conditions. The equations that make up the reward function of DQN are:

\begin{multline}
    R_1 = - \frac{1}{mn(n-1)}\sum^m_{z=0} \sum^n_{i=0} \sum^N_{j=0} SSIM_z(X_{i_z}, X_{j_z}) \\
    i \neq j\\
    \label{R1}
\end{multline}

\begin{multline}
    R_2 = - max_i(\frac{1}{m(n-1)}\sum^m_{z=0}\sum^n_{j=0} SSIM_z(X_{i_z}, X_{j_z}))\\
    i \neq j\\
    \label{R2}
\end{multline}
\begin{equation}
    R = R_1 + R_2 - n/3
    \label{R}
\end{equation}

Where $R_1$ represents the first reward, $R_2$ represents the second reward, $R$ represents the total reward, $m$ is the number of operating conditions, $n$ is the length of the dataset, and SSIM stands for structural similarity index measure \cite{wang2004image}. 

DQN uses a CNN framework that consists of five consecutive instances of a convolution layer, followed by a batch normalization layer, and activated by a ReLU activation function \cite{nair2010rectified}--the inputs are either Mel-spectrogram \cite{choi2017tutorial, yang2019machine} images of the sounds (discussed in Section \ref{datacollect}) or images of the system. The output of this CNN is flattened to create a vector representing the latent space of the input. This creates a single vector that goes through a single linear layer to reduce the dimensionality to ten or eight--the choice is problem specific. 

A softmax function activates the first eight outputs that tell the system where to collect data. The chosen location corresponds to the output with the highest value after activation. A separate softmax function activates the last two outputs to determine whether to use image or sound data. The last two outputs are not included if multimodal data is not required.

DQN runs for 100 episodes to learn how to collect the data optimally. The network uses a root mean squared propagation (RMSProp) optimizer \cite{hinton2012neural}. The data collected by DQN, $X^*$, will be used to train the FSL algorithm. 

\subsubsection{Few-Shot Learning}
The FSL algorithm used by HARDLearning is prototypical network introduced by Snell et al. \cite{snell2017prototypical}. Tangentially, FSL algorithms fall into two classes: data augmentation \cite{ sun2019meta, balaji2018metareg, yoo2017efficient, ziko2020laplacian} and architecture augmentation \cite{wang2020generalizing}. Prototypical network \cite{snell2017prototypical} is an architecture augmentation approach. Snell et al. \cite{snell2017prototypical} propose that there exists an embedding in which points cluster around a single prototype representation for each class. Prototypical network learns a non-linear mapping from the input into an embedding space and takes a prototype of a class to be the mean of its support set in the embedding space. Then, classification is performed for an embedded query point by finding the nearest class prototype. This technique is used over other FSL techniques because it does not increase the dataset size and relies solely on mapping to an embedding space that is beneficial for predicting with smaller datasets. The network used is modified slightly to handle multimodal inputs.

Prototypical network receives the dataset composed by DQN, and the network's structure and hyperparameters used in this paper are similar to the ones used by Snell et al. \cite{snell2017prototypical}; however, with multimodal inputs, we need to stack the inputs along their channels and create a single input with $3n$ channels. The CNN consists of four instances of a convolution layer, followed by a batch normalization layer, then activated by a ReLU activation function, and a down-sampling through a max-pooling layer using a 2$\times$2 kernel. The output of the CNN is similarly flattened, but the dimensionality is reduced through two linear layers. The dimensionality is reduced to a vector of length 128. A log-softmax activation function then activates the output vector to give a representation of all the classes.

The number of epochs used by prototypical network depends on the problem set, but 100 epochs were used for the fan problem, and 1000 epochs were used for the Navistar engine problem--which will be explained in Section \ref{ProbSet}. However, the optimizer and loss function remained constant; the optimizer used is an adam optimizer with $\beta_1 = 0.5$, $\beta_2 = 0.999$ and the learning rate is 0.0002 with a negative log-likelihood loss function \cite{yao2019negative}. After training, the network's weights are saved. 

From DQN, we receive the set of locations, $P^*$, and from prototypical network, we receive the trained model's weights; both are deployed in the HAR application to develop a testing dataset and perform fault diagnostics on the real-world testing data collected.
 
\subsubsection{Evaluation}
The rewards of DQN are used to see if a local minimum is found, and then we use the precision and recall (equations \ref{precision}-\ref{recall}) of prototypical network using different datasets to show the relationship between the rewards of the DQN and the success of prototypical network.

\begin{equation}
Precision = \frac{True\;Positive}{True\;Positive + False\;Positive}
\label{precision}
\end{equation}

\begin{equation}
Recall = \frac{True\;Positive}{True\;Positive + False\;Negative}
\label{recall}
\end{equation}

This completes the training part of the framework outlined in Figure \ref{fig:interactions_main}, but the human element has yet to be discussed.

\subsection{Deployment}
The deployment phase in the lower half of Figure \ref{fig:interactions_main} deploys the stored information from both DQN and prototypical network in a HAR environment--on a mobile application. Sahu et al. \cite{sahu2020artificial} motivate the use of HAR; the authors review the intersection of artificial intelligence and AR and conclude that the fusion of the two fields can assist in manufacturing applications. Our HAR environment interfaces between the human user and prototypical network. $P^*$ and the data types are returned to the user sequentially to provide instructions for collecting data similar to $X^*$. The instructions are deployed in a HAR application to improve clarity. 
The process of explaining to a user how to collect test data appends $S_2$ to $H_{AI}$, improving the human-AI communication pipeline for problem-solving. If the user collects the data properly, prototypical network will achieve the highest accuracy possible–validating the conclusions of Sahu et al. \cite{sahu2020artificial}. 

\begin{figure}[H]
    \centering
          \begin{subfigure}{0.15\textwidth}
        \includegraphics[width=\textwidth]{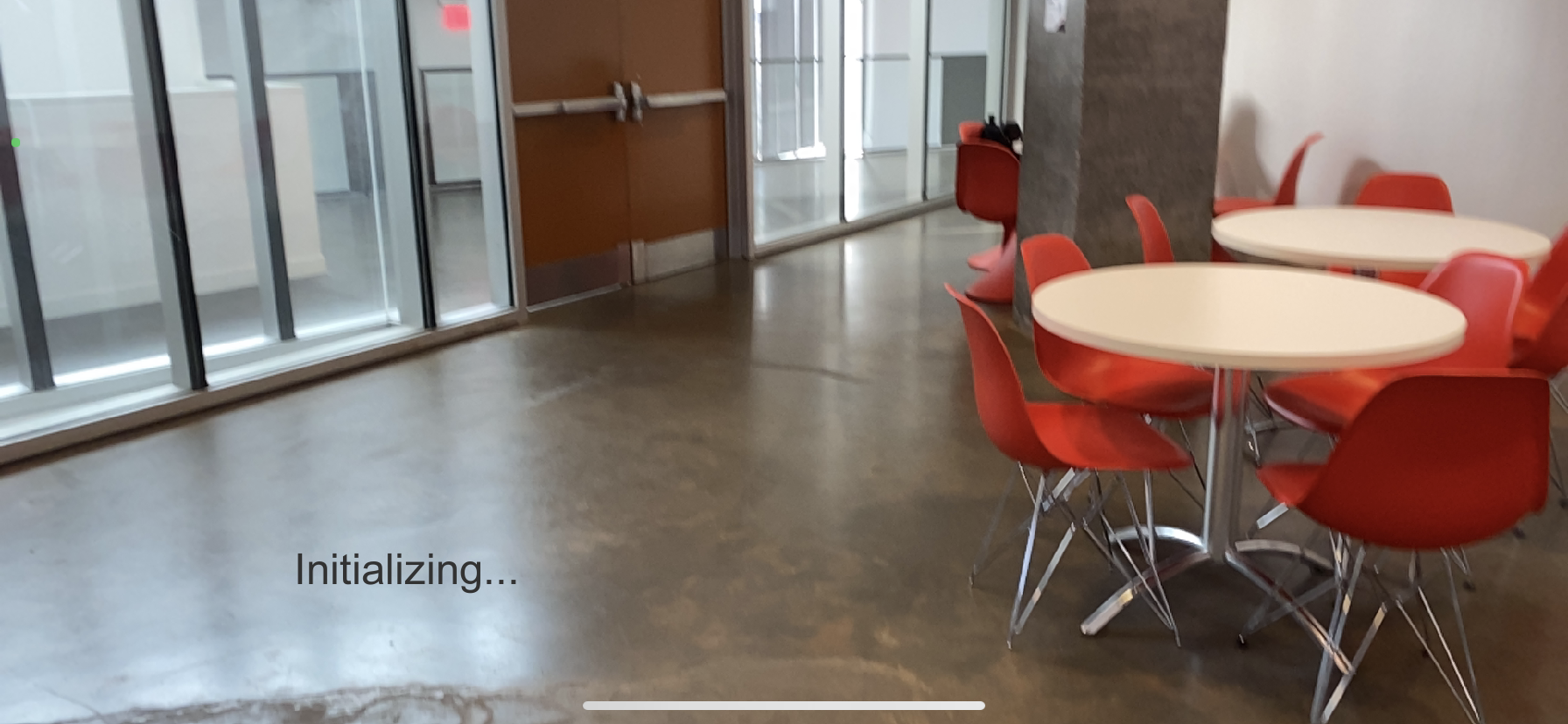}
          \caption{Step 1}
          \label{fig:Step1}
      \end{subfigure}
      \begin{subfigure}{0.15\textwidth}
        \includegraphics[width=\textwidth]{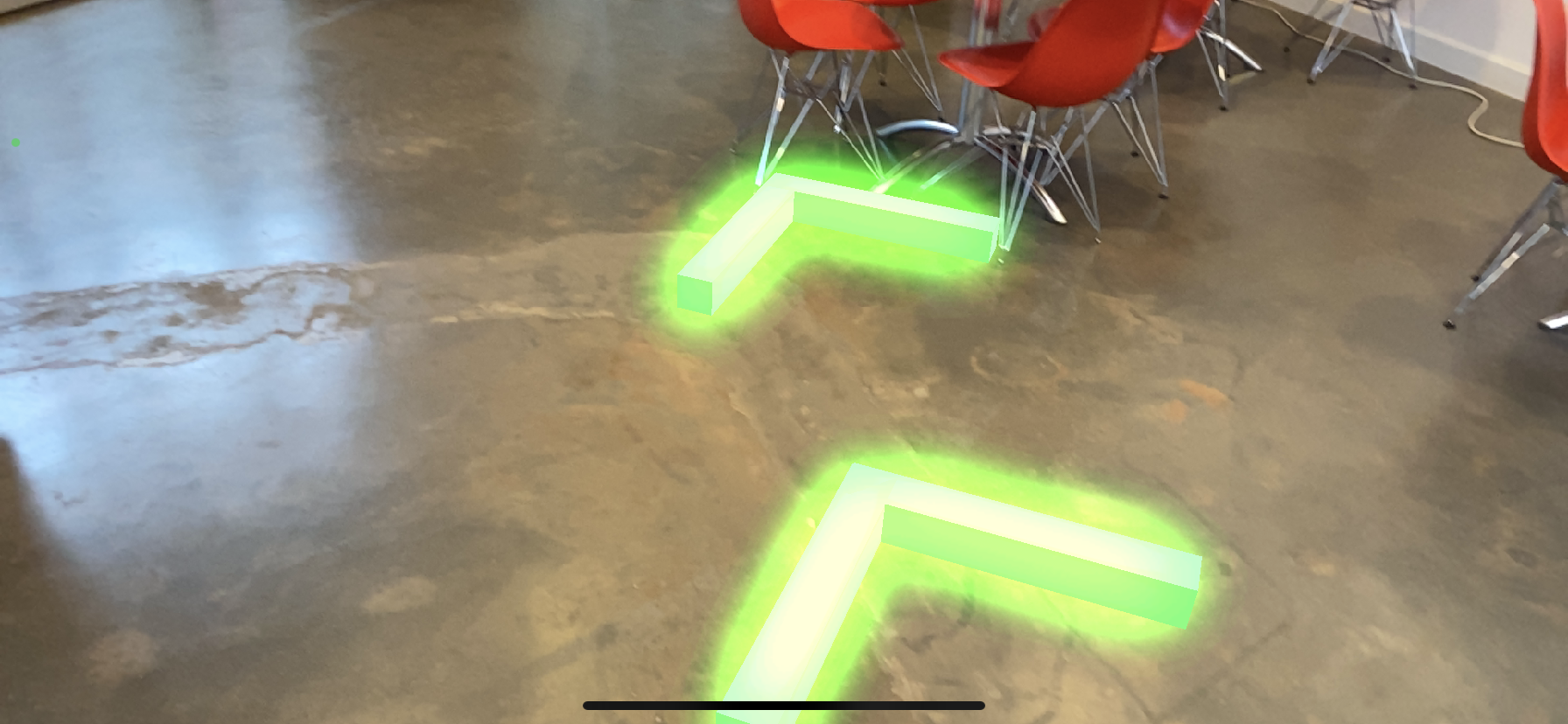}
          \caption{Step 2}
          \label{fig:Step2}
      \end{subfigure}
      \hfill
      \begin{subfigure}{0.15\textwidth}
        \includegraphics[width=\textwidth]{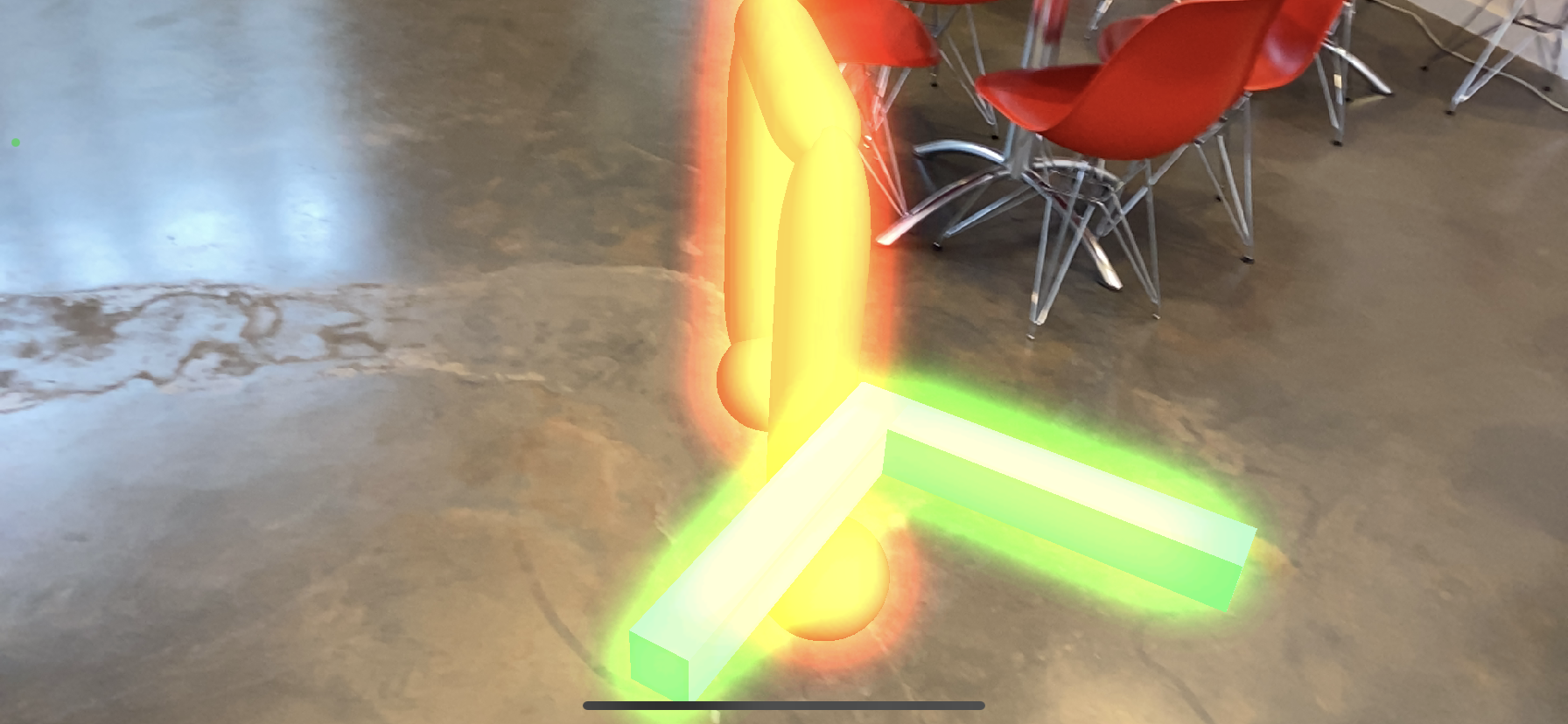}
          \caption{Step 3}
          \label{fig:Step3}
      \end{subfigure}
            \begin{subfigure}{0.15\textwidth}
        \includegraphics[width=\textwidth]{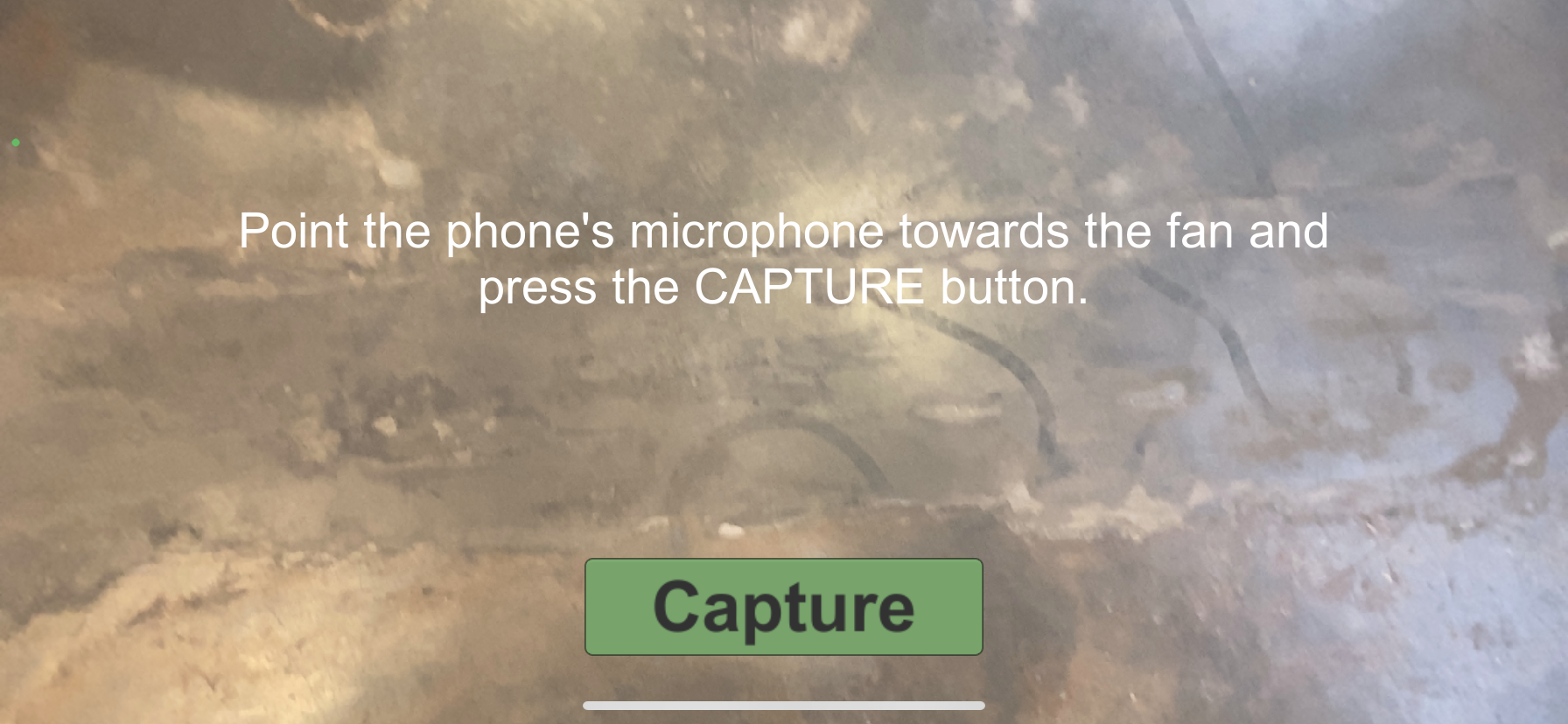}
          \caption{Step 4}
          \label{fig:Step4}
      \end{subfigure}
      \hfill
      \begin{subfigure}{0.15\textwidth}
        \includegraphics[width=\textwidth]{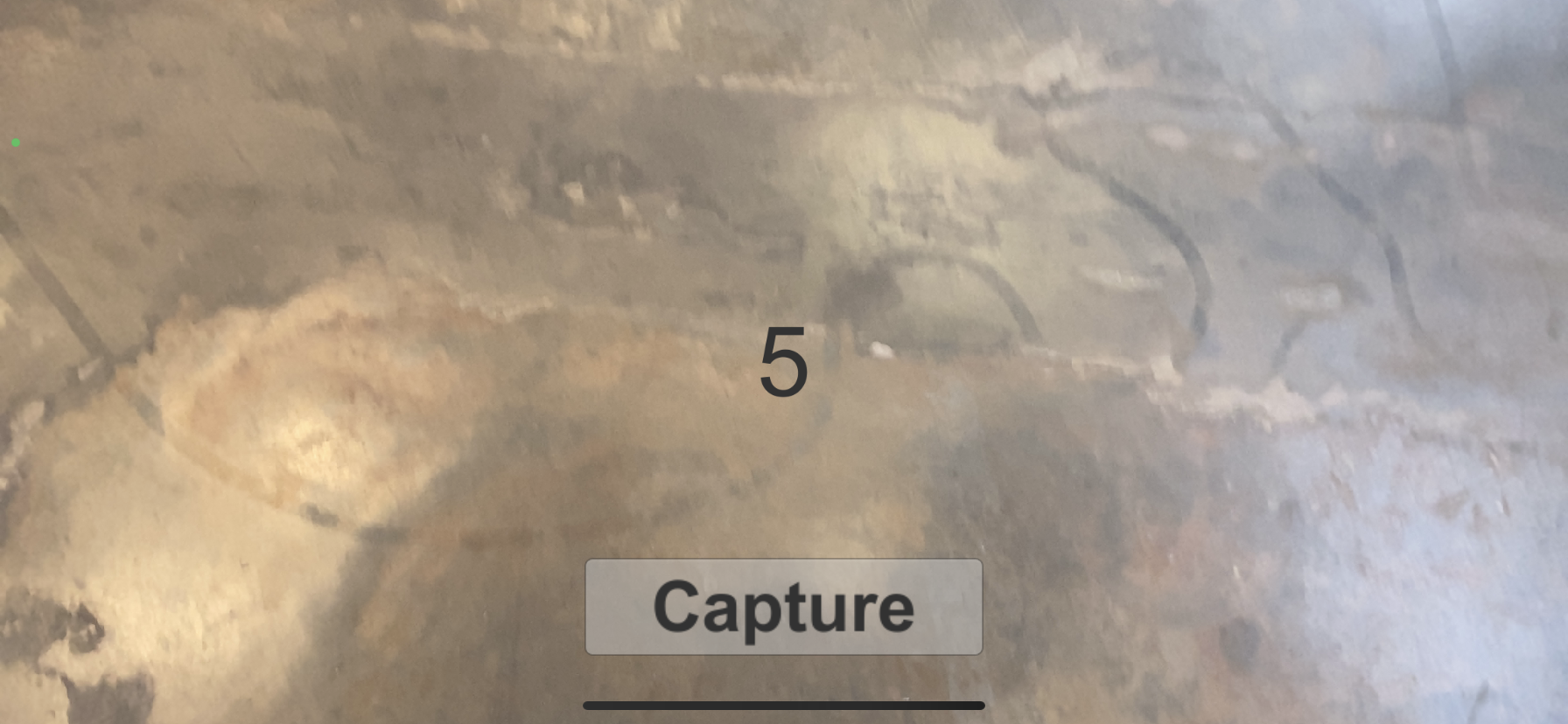}
          \caption{Step 5}
          \label{fig:Step5}
      \end{subfigure}
            \begin{subfigure}{0.15\textwidth}
        \includegraphics[width=\textwidth]{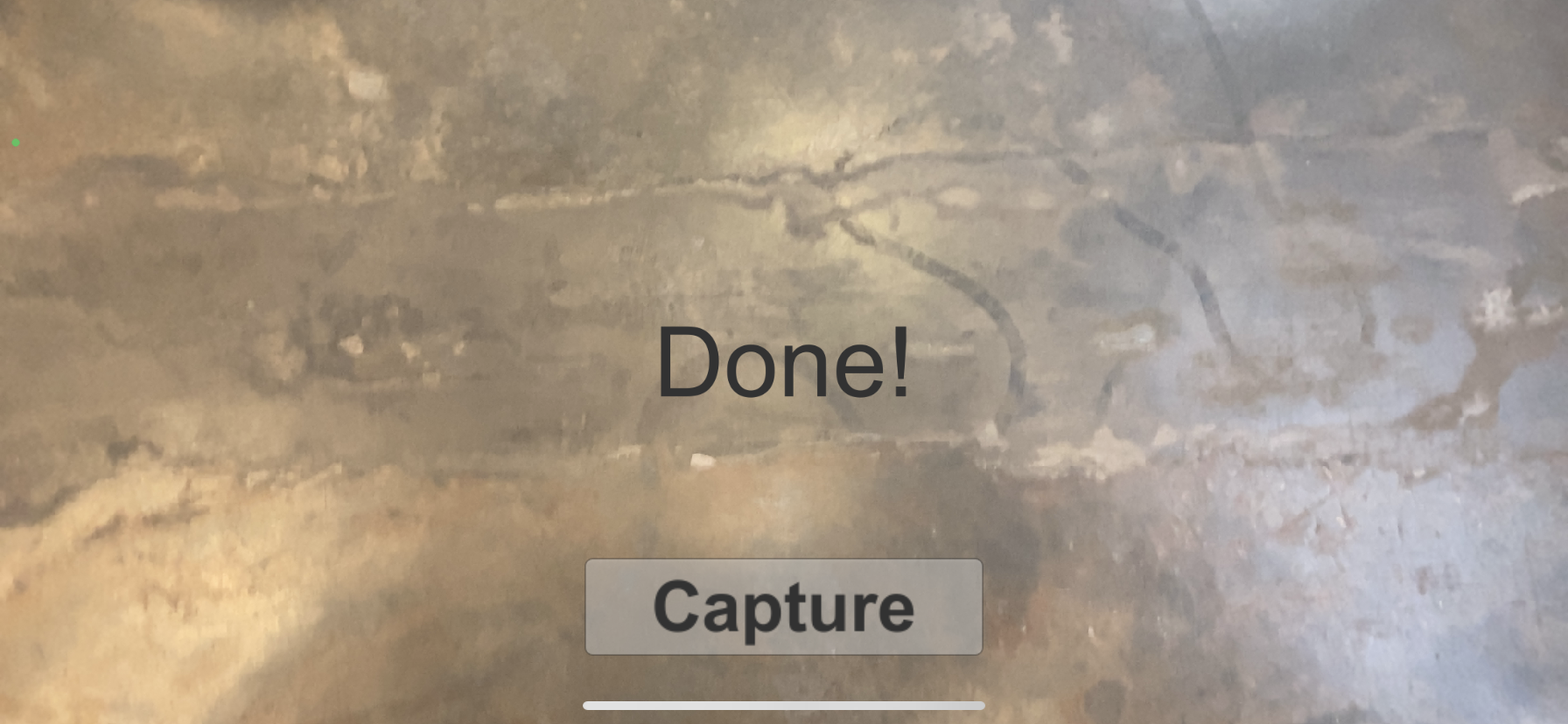}
          \caption{Step 6}
          \label{fig:Step6}
      \end{subfigure}
      \hfill
      \begin{subfigure}{0.15\textwidth}
        \includegraphics[width=\textwidth]{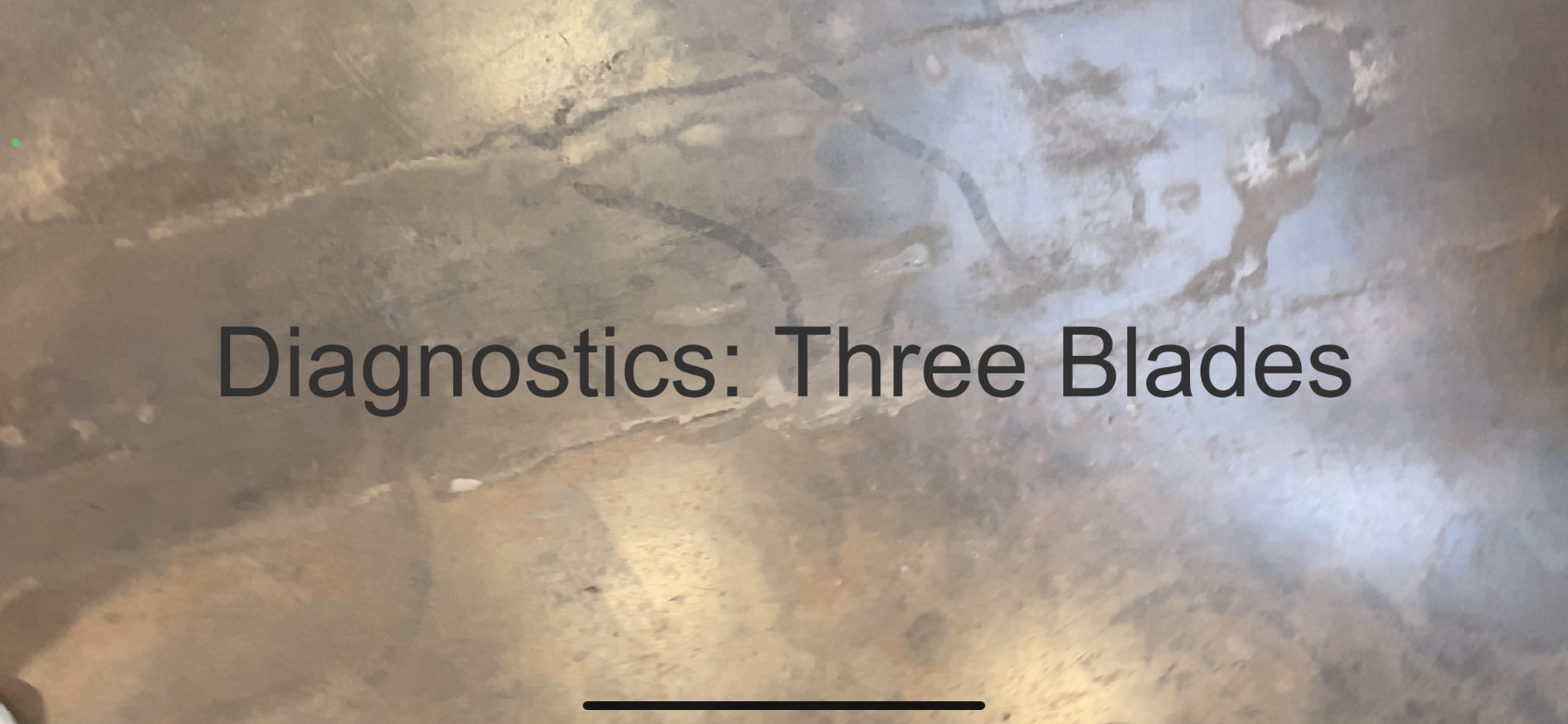}
          \caption{Step 7}
          \label{fig:Step7}
      \end{subfigure}
\caption{HARDLearning application instructions for data collection}
\label{fig:Steps}%
\end{figure}

$S_2$ is shown in Figures \ref{fig:Step1}-\ref{fig:Step7}. The first phase (steps 1-2) is initialization; the application needs to know where the user is on the map. The application uses local features to determine the user's position to populate the screen with directions. Next, the navigation phase (steps 2-3) shows the directions visually using waypoints that augment the camera view. As the system operator approaches the end of the navigation, a destination icon will appear, and this location is represented by either a camera. The camera signifies that the system operator should capture a picture at that location, and the music note signifies that they should record the audio.

The navigation to the capture location uses an $A^*$ algorithm \cite{hart1968formal}.

Once the system operator reaches the data capture location, the screen provides instruction, and the data collection phase begins (steps 4-7). The instruction will tell the operator to point the camera or microphone towards the system. Then after a few seconds, a button will populate the screen with the message: "Capture." The operator will click the button, and the capture will occur. Both options will collect information for the duration necessary, and since the audio recording lasts five seconds, it will count down from five. After the countdown is complete, a message telling the user that the capture is done will appear; finally, the diagnosis will appear once the trained prototypical network fully processes the data.

There should be similarities between the training and testing data by collecting the data as instructed. Minimizing this difference means that fitting to the training data will yield better results during testing; therefore, any overfitting caused by the model's training will have minimal effect on the efficacy of the prediction on the testing data. 

\subsubsection{Evaluation}
We evaluate the usability of the HAR application by acquiring volunteers to use the application and provide feedback. The volunteers are also timed on their experience using the application to have an objective metric for performance. Nielsen and Landauer \cite{nielsen1993mathematical} claim that the number of usability problems found, $F_P$, in a usability test with $u$ users follows the relationship in equation \ref{usability}.

\begin{equation}
    F_P=T_P(1-(1- L)^u).
    \label{usability}
\end{equation}
Where $T_P$ is the total number of problems and $L$ is the proportion of usability problems discovered while testing a single user, and Nielsen and Landauer \cite{nielsen1993mathematical} find $L$ is typically $31\%$. Additionally, they claim that with 15 users, all usability problems will be found, but five users can capture $85\%$ of the usability issues. We acquired 15 users, and after the seventh user, novel feedback was absent, driving us to believe that for this particular application, $L>31\%$. This is outlined further in Section \ref{Results}.



\section{Problem sets}
\label{ProbSet}
A novel diagnostics problem set is introduced to showcase the efficacy of HARDLearning in the fault diagnostics domain. The data collected to develop the problem set is not used to train prototypical network directly--it is used to develop a virtual environment for DQN to explore. The development of the virtual environment for each problem set and the preprocessing of the data are outlined below.
\subsection{Virtual Environment Development}
\label{datacollect}
\begin{figure}
\begin{subfigure}{.15\textwidth}
  \centering
  \includegraphics[width=.5\linewidth]{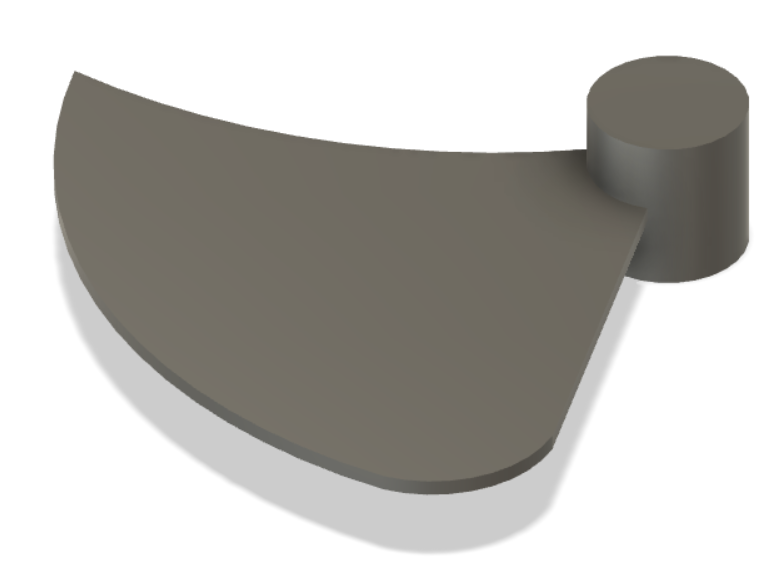}  
  \caption{1 Blade}
  \label{fig:1blade}
\end{subfigure}
\begin{subfigure}{.15\textwidth}
  \centering
  \includegraphics[width=.5\linewidth]{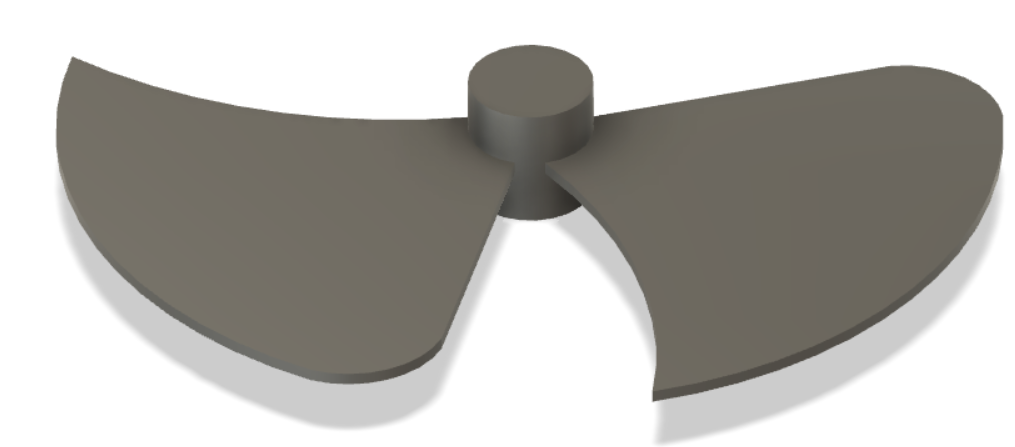}  
  \caption{2 Blades}
  \label{fig:2blades}
\end{subfigure}
\begin{subfigure}{.15\textwidth}
\centering
  \includegraphics[width=.5\linewidth]{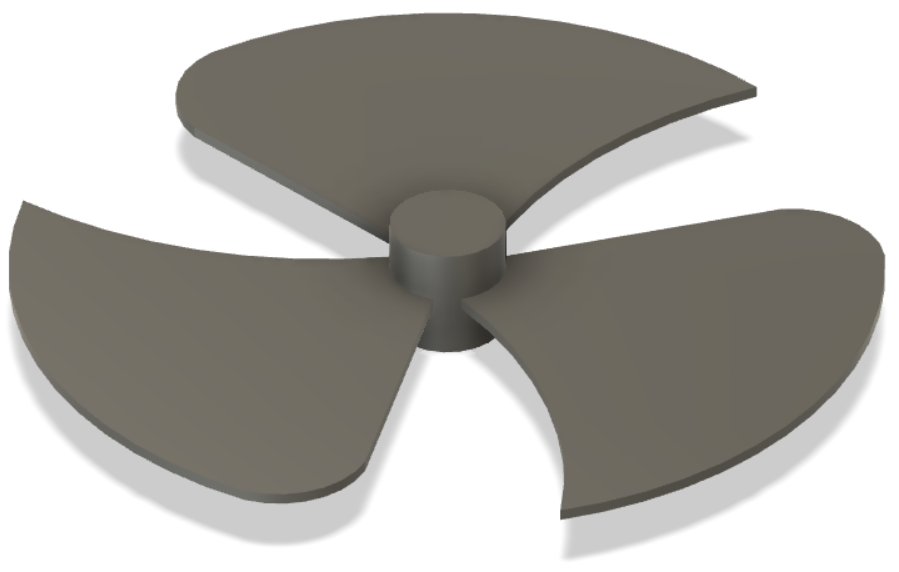}  
  \caption{3 Blades}
  \label{fig:3blades}
\end{subfigure}
\begin{subfigure}{.15\textwidth}
  \centering
  \includegraphics[width=.5\linewidth]{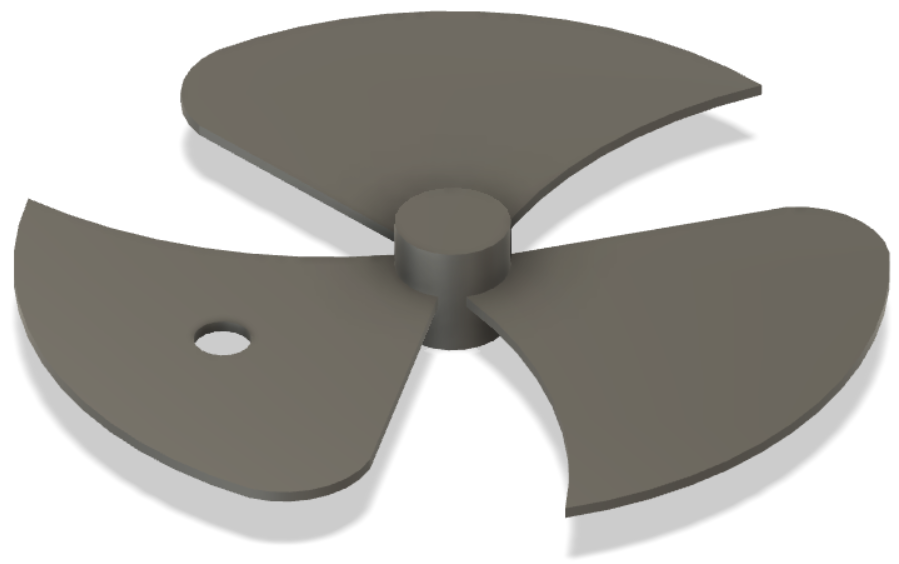}  
  \caption{1 Hole}
  \label{fig:1hole}
\end{subfigure}
\begin{subfigure}{.15\textwidth}
\centering
  \includegraphics[width=.5\linewidth]{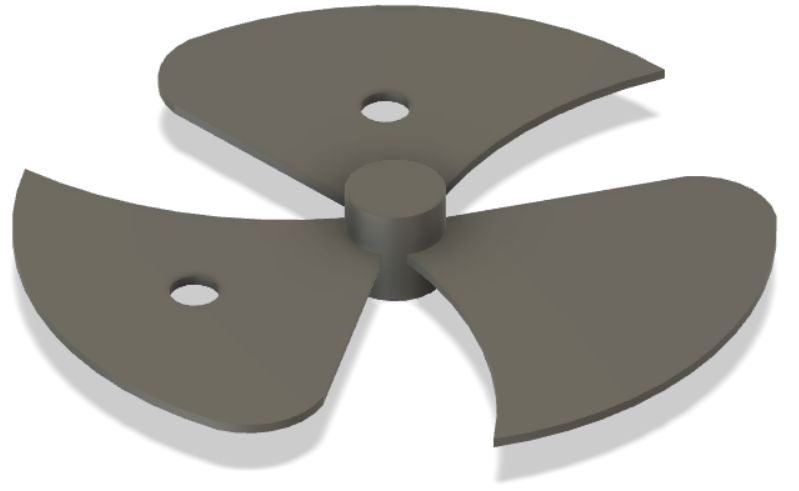}  
  \caption{2 Holes}
  \label{fig:2holes}
\end{subfigure}
\begin{subfigure}{.15\textwidth}
  \centering
  \includegraphics[width=.5\linewidth]{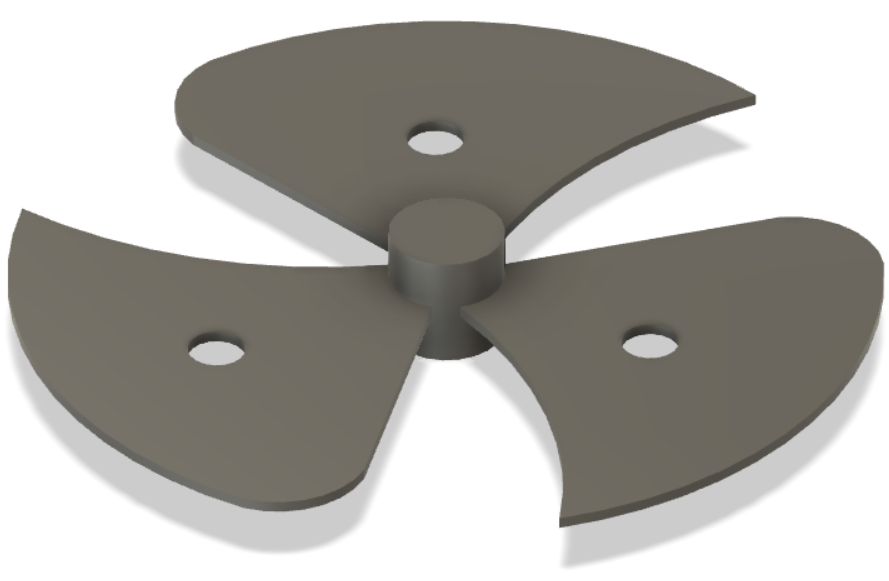}  
  \caption{3 Holes}
  \label{fig:3holes}
\end{subfigure}
\caption{3D models of the six different fan operating conditions}
\label{fig:3D}
\end{figure}

The fan problem set considers six operating conditions: one blade, two blades, three blades, one hole, two holes, and three holes (shown in Figures \ref{fig:1blade}-\ref{fig:3holes}). This environment will have visual and auditory modalities, so we outline the implementation of both. The visual modality is enabled by developing a 3D model of the fan in Blender; the actions available for this modality are rotating and translating the camera to capture images of each operating condition at the desired pose. 

The auditory modality was enabled by generating audio files in Reaper of the fan under each operating condition at various locations. The audio files were generated by developing a real-world system model and recording the fan's sound in operation at 0-degrees, 90-degrees, 180-degrees, and 270-degrees at distances of 1ft and 5ft. The 90-degree intervals were chosen as they provided the most extreme cases for both the sound and visual data while minimizing the search space. Two distances were chosen because there was minimal change in the sound between 1ft and 5ft; plus, 5ft maximized the distance available when capturing the data. At these locations, sounds were recorded for five seconds. Longer recordings will yield a more significant difference between operating conditions but will be prone to more error and make the feature extraction more difficult for the learning algorithm. Yang and Rai \cite{yang2019machine} use 0.2 seconds of data to develop their sound profiles to make feature extraction easier; however, some features are not represented in 0.2 seconds for the data collected, which weakens the performance of the learning algorithm. We increased the recording time until the SSIM between operating conditions stopped decreasing, occurring at around five seconds. Therefore, five seconds of audio helped maximize the difference between operating conditions while minimizing the error propagation and the difficulty of feature extraction. 

Since there are only eight locations where auditory data exists, the actions for this modality include rotating and translating amongst the eight locations. Subsequently, we constrain the visual modality search space to the same eight locations. 

\subsection{Preprocessing}
The sampled interval for each audio recording was transformed into Mel-spectrogram images. To create a Mel-spectrogram image, fast Fourier Transform (FFT) first converts waveform data in the time domain into a frequency domain. Then, a spectrogram image is obtained after extracting the amplitude-frequency information of the waveform data in the frequency domain. The spectrogram image plots frequency against time. After developing a spectrogram image, the frequency is expressed in Mel scale to create a Mel-spectrogram image. The frequency ($f$) in hertz is converted into Mels ($m$) using equation \ref{eq:mel}.

\begin{equation}
\label{eq:mel}
m = 2595\log_{10}(1 + \frac{f}{700})
\end{equation}

\begin{figure}[H]
\centering
\begin{subfigure}{.2\textwidth}
  \centering
  \includegraphics[width=0.5\linewidth]{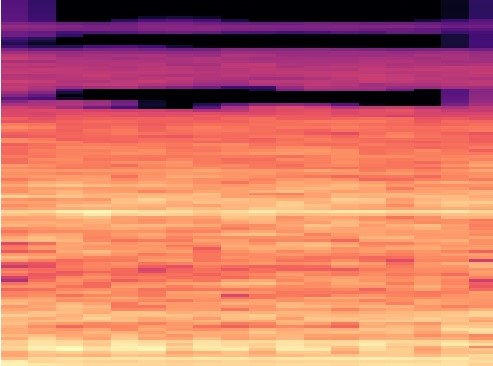}  
\end{subfigure}
\begin{subfigure}{.2\textwidth}
  \centering
  \includegraphics[width=0.5\linewidth]{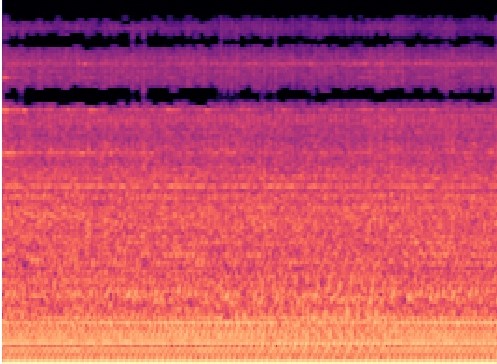}  
\end{subfigure}
\begin{subfigure}{.2\textwidth}
\centering
  \includegraphics[width=0.5\linewidth]{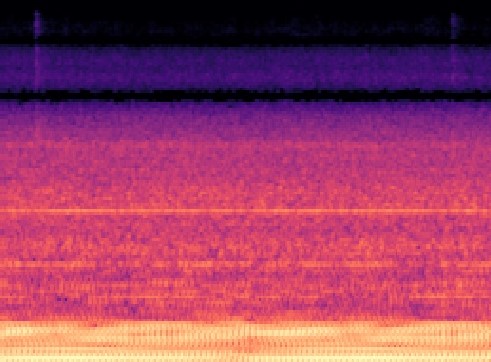}  
\end{subfigure}
\begin{subfigure}{.2\textwidth}
  \centering
  \includegraphics[width=0.5\linewidth]{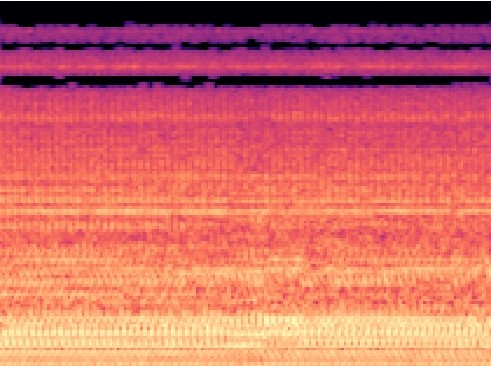}  
\end{subfigure}
    \caption{Example Mel-Spectrograms of the sound collected}
    \label{fig:sound_ex}
\end{figure}

These plots are saved as images. Examples of Mel-spectrogram images are shown in Figure \ref{fig:sound_ex}.

All images captured have dimensions: 480 pixels by 640 pixels (480$\times$640), but the images are resized to 120$\times $160 to reduce the size of the network. The image is normalized by dividing each element in the matrix by 255. All of the images captured are compiled into a list of tensors, and the data is organized in the following way: (1) the index of the image in the list, (2) the channels of color, (3) the height, and (4) the width.


\section{Results}
\label{Results}

This section will evaluate each phase of the adapted $H_{AI}$ framework. The final locations chosen for data collection will be evaluated to measure the efficacy of $S_0$, the results of prototypical network will reflect the quality of $S_1$ for generating $F$, and the HAR application will represent $S_2$--and the usability study will benchmark the performance.

After training the HARDLearning pipeline, the algorithm converges to collecting sound at the 5ft 0-degrees location. The trained model showed precision and recall of $100\%$ on the training set provided (Table \ref{tab:P_R_optimal1}), demonstrating the architecture's efficacy on a simple problem. We explore the data collection process and the efficacy of prototypical network on the data collected to explain the results.
\begin{table}[H]
    \centering
    \begin{tabular}{|c|c|c|c|}
    \hline
         
         Location & Precision & Recall  \\
         \hline
         5ft 0-degree & 1 & 1\\
          \hline
    \end{tabular}
    \caption{Fan Problem: Precision and Recall of Optimal Data}
    \label{tab:P_R_optimal1}
\end{table}

\subsection{Data Collection}



Table \ref{tab:SSIM} outlines the SSIM across the operating conditions for each positional configuration. The SSIM values at different configurations are used to show how the data collected affects the precision and recall in Table \ref{tab:P_R}. The lowest similarity from the visual data appears at the 0-degree and 180-degree locations--there is a $7.76\%$ (1ft) and $4.75\%$ (5ft) decrease in SSIM from the 270-degree locations. The 90-degree and 270-degree locations hide the features that differentiate the operating conditions. There is also a decrease in similarity when the camera is closer to the fan. Similarly, this was expected because the differentiating features of the operating conditions occupy more pixels than when the camera is further. However, the auditory data follows a different trend; the lowest similarities occur at the 0-degree locations. The 0-degree locations showed a $4.95\%$ (1ft)  and $6.32\%$ (5ft) decrease in SSIM from the 180-degree locations. Since the fan is projecting air forward, the sound behind the fan has a greater similarity between operating conditions. The distance between the fan and the microphone produced a negligible change in the SSIM; further distances could be explored to determine the effect on SSIM from distance from the fan. Nevertheless, we can conclude that the change in the distance had minimal effect on the SSIM.

After 100 episodes, DQN converges to collecting auditory data at the 5ft 0-degrees location. This conclusion is consistent with our data because that location has the lowest SSIM, which would yield the highest reward. After it converges to this location, it maintains this position. Our reward function helps DQN  find the global minimum.

\begin{table}[H]
    \centering
    \begin{tabular}{|c|c|c|c|}
    \hline
         Distance & Angle & Image & Sound  \\
         \hline
         \multirow{4}{*}{1ft} & 0 & 0.8724 & 0.7694 \\
         \cline{2-4}
         & 90 & 0.9399 & 0.7730\\
         \cline{2-4}
         & 180 & 0.8724 & 0.8095\\
         \cline{2-4}
         & 270 & 0.9458 & 0.7704\\
          \hline
        \multirow{4}{*}{5ft} & 0 & 0.9458 & 0.7653 \\
         \cline{2-4}
         & 90 & 0.9928 & 0.7678\\
         \cline{2-4}
         & 180 & 0.9458 & 0.8169\\
         \cline{2-4}
         & 270 & 0.9930 & 0.7721 \\
          \hline
    \end{tabular}
    \caption{Fan Problem: SSIM across operating conditions with varying configurations}
    \label{tab:SSIM}
\end{table}

\subsection{Deep Learning Model}

Following the results reported about the data collection, Table \ref{tab:P_R} captures prototypical network's ability to differentiate between operating conditions given different information types. The precision and recall are averaged over the six operating conditions, and for nearly all locations, auditory data provides better precision and recall after training prototypical network. This observation aligns with our hypothesis that lower similarity between operating conditions will increase the efficacy of the DL; the auditory data had lower SSIM values than visual data universally. Secondly, this validates our reward function that minimizes SSIM over the operating conditions. Predictably, our reward function is not perfect because even though the lower SSIM typically yields a better result, there are a few exceptions to this rule. Therefore, more variables must control the accuracy of the prototypical network. Future work will explore reward functions that represent these details to contribute to the success of training prototypical network. Nevertheless, the chosen location has a precision and recall of $1$ and $1$, respectively; and shows a $700\%$ and $200\%$ improvement from training on the worst data.

\begin{table}[H]
    \centering
    \begin{tabular}{|c|c|c|c|c|}
    \hline
         Distance & Angle & Metric & Image & Sound  \\
         \hline
         \multirow{4}{*}{1ft} & \multirow{2}{*}{0} & Precision & 0.5 & 0.75\\
         \cline{3-5}
         & & Recall & 0.667 & 0.833 \\
         \cline{2-5}
         & \multirow{2}{*}{90} & Precision & 0.375 & 0.5 \\
         \cline{3-5}
         & & Recall & 0.5 & 0.667 \\
         \cline{2-5}
         & \multirow{2}{*}{180} & Precision & 0.75 & 0.75\\
         \cline{3-5}
         & & Recall & 0.833 & 0.833 \\
         \cline{2-5}
         & \multirow{2}{*}{270} & Precision & 0.375 & 0.139\\
         \cline{3-5}
         & & Recall & 0.5 & 0.333\\
          \hline
        \multirow{4}{*}{5ft} & \multirow{2}{*}{0} & Precision & 0.125 & 1\\
         \cline{3-5}
         & & Recall & 0.333 & 1\\
         \cline{2-5}
         & \multirow{2}{*}{90} & Precision & 0.375 & 0.75\\
         \cline{3-5}
         & & Recall & 0.5 & 0.833 \\
         \cline{2-5}
         & \multirow{2}{*}{180} & Precision & 0.125 & 0.333  \\
         \cline{3-5}
         & & Recall & 0.333 & 0.5 \\
         \cline{2-5}
         & \multirow{2}{*}{270} & Precision & 0.375 & 0.75 \\
         \cline{3-5}
         & & Recall & 0.5 & 0.833\\
          \hline
    \end{tabular}
    \caption{Fan Problem: Precision and Recall with varying configurations}
    \label{tab:P_R}
\end{table}


\subsection{Testing and Usability}
\subsubsection{Testing}
\begin{figure}[H]
\begin{subfigure}{.22\textwidth}
  \centering
  \includegraphics[width=0.5\linewidth]{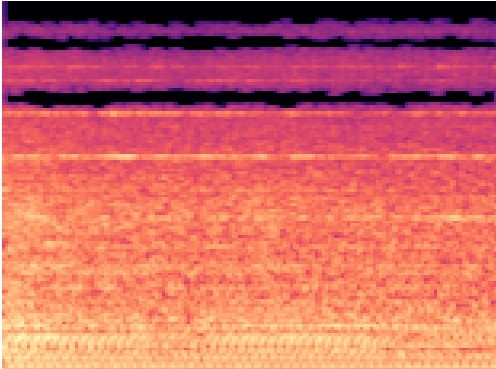}  
\end{subfigure}\hfill
\begin{subfigure}{.22\textwidth}
  \centering
  \includegraphics[width=0.5\linewidth]{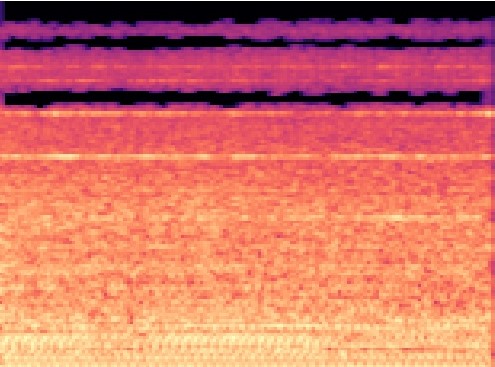}  
\end{subfigure}\hfill
    \caption{Mel-Spectrograms of the 1-blade fan sound data collected at the 5ft 0-degree location for training (left) and 5ft 0-degree location  for testing (right).}
    \label{fig:Fa_sounds}
\end{figure}

After training, the instructions for collecting the data and the trained few-shot learning model are deployed on a mobile device. Then, a user is instructed to collect the required data for diagnostics. We used the HARDLeaning app to navigate to the desired location(s) and collect the required data for all different operating conditions. The collection process was repeated five times for all operating conditions belonging to each dataset. Figure \ref{fig:Fa_sounds} give examples of the data collected during training and testing for a visual comparison of the collected data.

Additionally, the SSIM between the training and testing data collected was measured to show how well the trained prototypical network handled differences in the training and testing data. For the fan dataset, the average SSIM between the training and testing data for each operating condition was $0.9612\pm0.0105$. Furthermore, the precision and recall were unfazed by the difference in the data collected as the results were the same. The precision and recall for the fan dataset were $1$ and $1$, respectively. As long as the data collected by the user has the highest similarity to the corresponding class in the training data, prototypical network should categorize the operating condition properly. By following the instructions in the application, a user should be providing the trained prototypical network with data highly similar to the data it was trained on. Therefore, reducing the issue of overfitting for FSL and maximizing the model's accuracy.

\subsubsection{Usability}
After evaluating the applicability to real-world data, we explore the usability of the software. Importantly, all users were able to complete the initialization, navigation, and data collection phase without additional guidance. However, usability problems increased the time spent using the application, which should be addressed in future work.

\textbf{Initialization.}
The initialization steps are step 1 and step 2 in Figures \ref{fig:Step1}-\ref{fig:Step2}. The quantitative results from the usability trials show that it takes users on average $14.50\pm12.45$ seconds to complete this step, which is an $8.20$ second increase from the $6.30\pm0.30$ seconds it took the authors to complete this step. This part requires users to scan the environment to locate themselves on the map. Each user has a different scanning speed, which adds variance to the initialization time; however, such an increase could also indicate usability problems at this step. Looking at the qualitative results adds clarity to this sharp increase in time. This portion received the lowest subjective score at $3.93$ out of a maximum score of $5$. Users mentioned that they were unsure if they were scanning the environment properly and suggested that the initialization screen provide instructions on how to initialize the environment optimally, such as: (1) provide what objects need to be scanned, (2) explain how long to scan each object, and (3) whether to stand in one place or walk around while scanning.

\textbf{Navigation.} The navigation steps are steps 2-3 shown in Figures  \ref{fig:Step2}-\ref{fig:Step3}.  The quantitative results from the usability trials show that it takes users on average $21.2\pm11.81$ seconds to complete this step, which is a $13.68$ second increase from the $7.52\pm0.44$ seconds it took the authors to complete this step. Contrastingly, the subjective score was $4.33$, indicating it was more user-friendly than initialization, yet the time required to complete this step increased more than the initialization step. Additionally, the qualitative results for this phase only reveal small inconveniences: (1) the arrows look like brackets, or (2) in some cases; the participants were unsure immediately whether to follow the arrows or click on the arrows. However, feedback given for the initialization phase reveals the increase in time. Some participants did not realize the initialization was completed; these participants continued to scan the environment even though the arrows had appeared, resulting in longer navigation to the desired location. This problem can be alleviated by adding a confirmation text telling users initialization has finished along with instructions to guarantee users are looking in the direction where the arrows have appeared and, if not, instruct the users on how to rotate towards the arrows.

\textbf{Data Collection.}  The data collection steps are steps 4-7 shown in Figures \ref{fig:Step4}-\ref{fig:Step7}.  The quantitative results from the usability trials show that it takes users on average $16.44\pm4.95$ seconds to complete this step, which is a $2.11$ second increase from the $14.33\pm2.24$ seconds it took the authors to complete this step. This step has $6.21$ seconds of downtime, where the user is just waiting for the app to collect data, so the remaining time is allocated towards reading the instructions, rotating the device in the proper orientation, and clicking the capture button. Users were most satisfied by this part of the application giving it an average subjective score of $4.53$. The only concerns were that one user found it difficult to read the text and another user was unsure where the microphone was located on the device. These suggestions could be alleviated by adding a background for the text to guarantee readability regardless of the environment. Then, we could include an arrow pointing to the microphone on the device. The user who could not find the microphone completed the task in $19.02$ seconds, so there was added time, but on average, the increase in time comes from the reading speed and achieving the correct orientation. The new users do not have the prior knowledge that the authors have about what the message says and what to do with the device to complete this step more quickly.

\section{Conclusion}
\label{Conclusion}

In this paper, we outlined a novel human-machine interaction framework: HARDLearning--HARDLearning effectively diagnosed failures in fans. The precision and recall of the six different modes for the fan were both $100\%$. These results improved the precision and recall from blindly providing data to the prototypical network, further motivating the necessity for providing useful data to DL algorithms. The results remained the same when the data was collected by following the instructions in the HARDLearning application, demonstrating that prototypical network could handle the differences in the training and testing data to make an accurate prediction. Additionally, the training and testing data was, on average, above $95\%$ similar (according to SSIM) for each dataset. 

Secondly, the instructions provided by the HAR interface were able to navigate 15 users to the desired location for data collection, and all users were able to follow the data collection instructions provided. For an improved user experience, more instructions for initializing oneself in the environment can be added. Additionally, providing users with notifications within the app that signal the end of one step and the beginning of the next would be useful for users wasting time trying to complete a step that was already completed. These improvements could reduce the time of usability and provide users with a diagnosis more quickly.

Finally, future work should develop a more comprehensive list of actions so that the DQN will have a larger search space for the optimal collection method. 
\section{Acknowledgement}
This work was supported by the Naval Surface Warfare Center ($\bold{NSWC}$) funding with Agreement No: GRANT 13234068.
\bibliography{References}
\bibliographystyle{IEEEtran}

\end{document}